\documentclass{article}

\usepackage[final, nonatbib]{neurips_2024}

\usepackage[utf8]{inputenc} % allow utf-8 input
\usepackage[T1]{fontenc}    % use 8-bit T1 fonts
\usepackage{hyperref}       % hyperlinks
\usepackage{url}            % simple URL typesetting
\usepackage{booktabs}       % professional-quality tables
\usepackage{amsfonts}       % blackboard math symbols
\usepackage{nicefrac}       % compact symbols for 1/2, etc.
\usepackage{microtype}      % microtypography
\usepackage{xcolor}         % colors
\usepackage{makecell}       % line break in tables

\usepackage{enumitem}       % to get rid of spacing in itemization
\usepackage{svg}
\usepackage{subcaption}
\usepackage{tikz}
\usetikzlibrary{positioning}
\usepackage{comment}

\newcommand{\cknn}{KNN}

\newcommand{\clp}{LP}
\newcommand{\nm}{Macenko}
\newcommand{\nr}{Reinhard}
\newcommand{\dl}{TCGA-LUSC-5}
\newcommand{\dc}{CAMELYON16}

\title{Do Histopathological Foundation Models Eliminate Batch Effects? A Comparative Study}

\author{%
  Jonah K\"omen \\
  Machine Learning Group\\
Technische Universit\"at Berlin \\
BIFOLD\thanks{Berlin Institute for the Foundations of Learning and Data, Berlin, Germany.} \\
Berlin, Germany \\
  \And
  Hannah Marienwald \\
  Machine Learning Group\\
Technische Universit\"at Berlin \\
BIFOLD\textsuperscript{$*$} \\
Berlin, Germany \\
  \AND
  Jonas Dippel \\
Machine Learning Group\\
Technische Universit\"at Berlin \\
BIFOLD\textsuperscript{$*$}  \\
Aignostics \\
Berlin, Germany \\
  \And
  Julius Hense \\
  Machine Learning Group\\
Technische Universit\"at Berlin \\
BIFOLD\textsuperscript{$*$} \\
Berlin, Germany \\
}

\begin{document}

\maketitle

\begin{abstract}
Deep learning has led to remarkable advancements in computational histopathology, e.g., in diagnostics, biomarker prediction, and outcome prognosis. Yet, the lack of annotated data and the impact of batch effects, e.g., systematic technical data differences across hospitals, hamper model robustness and generalization. Recent histopathological foundation models --- pretrained on millions to billions of images --- have been reported to improve generalization performances on various downstream tasks. However, it has not been systematically assessed whether they fully eliminate batch effects. In this study, we empirically show that the feature embeddings of the foundation models still contain distinct hospital signatures that can lead to biased predictions and misclassifications. We further find that the signatures are not removed by stain normalization methods, dominate distances in feature space, and are evident across various principal components. Our work provides a novel perspective on the evaluation of medical foundation models, paving the way for more robust pretraining strategies and downstream predictors.
\end{abstract}

\section{Introduction}\label{sec:introduction}

Increasingly complex tasks in digital histopathology are solved by deep learning (DL) models, e.g., disease diagnosis or subtyping \cite{bejnordi2017camelyon, coudray2018nsclc, campanella2019clinical, strom2020prostate, lu2021clam, lu2021origin, dippel2024ad}, prediction of clinical or molecular biomarkers \cite{kather2020pan, binder2021morphological, echle2021biomarkers, arslan2024systematic}, and outcome prognosis \cite{courtiol2019mesothelioma, skrede2020outcome, saillard2020survival, chen2022multimodal}. While the performance of supervised end-to-end models is limited by the lack of labeled training data, self-supervised learning is currently adopted as a promising solution \cite{ciga2021self, chen2022scaling, wang2023retccl, wang2022transformer, kang2023benchmarking, phikon, kaiko2024, xu2024gigapath, chen2024uni, vorontsov2024virchow_natmed, Dippel2024RudolfV}. Pretrained a plethora of unlabeled histopathology images, foundation models extract features that can be utilized to achieve excellent prediction performances for a wide variety of downstream tasks, outperforming models trained from scratch or pretrained on natural images \cite{kang2023benchmarking}.

However, previous research \cite{Howard2021} revealed significant differences in digital whole-slide images (WSIs) across tissue source sites (TSS), i.e., hospitals, labs, or biobanks. These disparities stem from differences in fixation and staining protocols, WSI scanners, and patient demographics, among others. Such site-specific peculiarities, also called site signatures, may correlate with labels in downstream tasks, which creates potential batch effects. It was shown that DL models directly trained on WSI patches indeed employ such batch effects for solving supervised learning tasks, leading to biased predictions and compromised generalizability across tissue source sites. This lack of robustness is potentially dangerous in high-risk diagnostic systems arising in digital histopathology.

\begin{figure}[t!]
    \centering
    \includegraphics[width=0.9\linewidth]{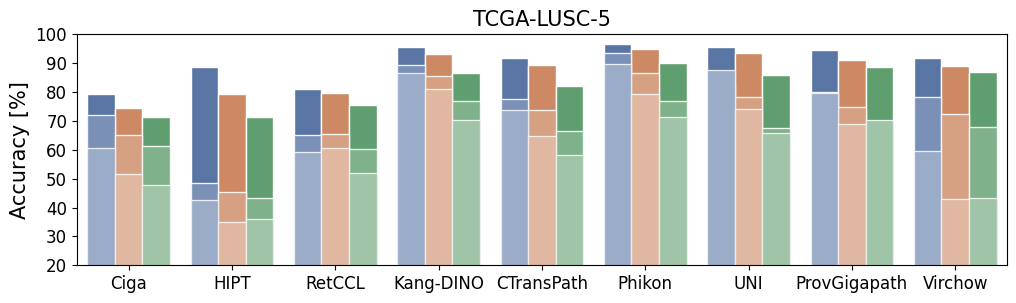} \\
    \includegraphics[width=0.9\linewidth]{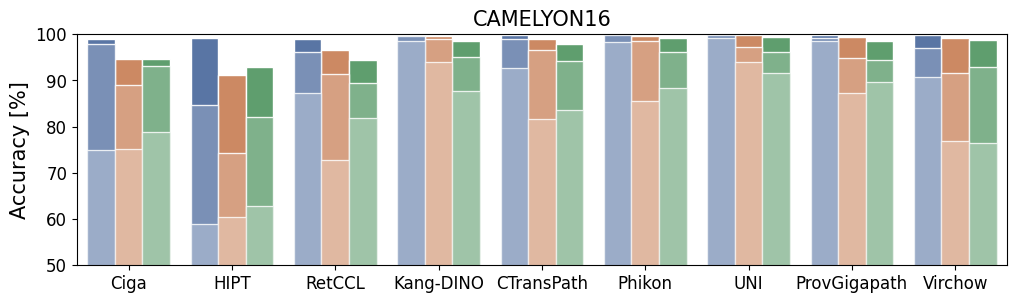} \\
    \vspace{1mm}
    \includegraphics[width=0.8\linewidth]{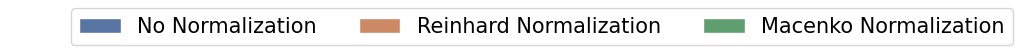}
    \caption{Accuracy scores for predicting the TSS of a patch from the feature embedding of different histopathological foundation models. We considered features of raw and stain-normalized patches \cite{reinhard, macenko} from two datasets: \dl{} (top) with 5 and \dc{} (bottom) with 2 tissue source sites. The most transparent bars represent the nearest centroid classifier, the medium transparent k-nearest neighbors, and the non-transparent bars linear probing models. \textit{Higher accuracy indicates stronger site signatures.} The experimental details are described in Section \ref{sec:source-site-prediction}.}
    \label{fig:source-site-prediction}
\end{figure}

As histopathological foundation models are trained on an excess of diverse multi-site data sets, employ augmentation strategies during pretraining, and have been reported to generalize better to unseen data \cite{chen2024uni, vorontsov2024virchow_natmed, Dippel2024RudolfV, campanella2024benchmark}, one may hypothesize that they are fully immune to data biases.
In this work, we show that they are in fact \textit{not}: all tested models are still susceptible to batch effects.
Our key findings include:
\begin{itemize}[noitemsep,topsep=0pt]
\item The feature embeddings of all foundation models contain site-specific signatures that can be accurately retrieved using linear methods. Interestingly, we observe the best-performing models according to the literature to exhibit the highest source site prediction accuracy. Figure \ref{fig:source-site-prediction} shows a preview of this result.
\item Site-specific information acts as a batch effect. Correlation between site signatures and downstream task labels may result in biased predictions and misclassifications. Commonly applied stain normalization techniques like \nr~\cite{reinhard} and \nm{}~\cite{macenko} do not sufficiently reduce such biases.
\item Distances between patches in feature space are more affected by discrepancies from the TSS rather than biological differences, which renders downstream algorithms resorting to the proximity of points, such as clustering, nearest neighbor search, or similar, as ill-posed.
\item The linear directions of the largest variance of the features exhibit high site separability, confirming that batch effects are very accessible, easily misleading supervised downstream algorithms.
\item Mitigating batch effects in feature space or during the training of foundation models is a promising research direction for improving the robustness and generalization of downstream predictors.
\end{itemize}

Our findings are based on several experiments that we describe in the remainder of the paper. After discussing relevant related work in Section~\ref{sec:related_work}, we provide a brief overview of the tested foundation models and the datasets we employ in Section~\ref{sec:models_data}. In Section~\ref{sec:identification}, we expose batch effects in histopathological foundation models by identifying that the issuing site can be easily predicted from feature vectors and that this biases downstream tasks. We further characterize this batch effect by examining distances between patches in feature space and relating feature variance with site separability in Section~\ref{sec:characterization}.

% TODO delete
%\newpage

\section{Related work}\label{sec:related_work}

Previous studies showed that supervised models are confounded by staining, scanner, and site-specific signatures \cite{Howard2021, histo-xai-review}. When the labels are correlated with these metadata, models possibly exploit them as shortcut/Clever Hans features, leading to catastrophic failure on new uncorrelated cases \cite{clever-hans, geirhos2020shortcut, hermannfoundations, unsupervised-clever-hans}.

Self-supervised learning models, trained on an abundance of diverse unlabeled data with different metadata, are proposed as a solution and observed to be more robust in some domains \cite{navarro2022self, goyal-ssl-robust}. In histopathology, scaling dataset and model sizes recently led to significant performance improvements on downstream tasks \cite{chen2024uni, vorontsov2024virchow_natmed, xu2024gigapath, Dippel2024RudolfV}. However, to the best of our knowledge, it has not been investigated whether this also results in foundation models that eliminate batch effects. Existing works on evaluating histopathological foundation models focused on performance assessments on a variety of (clinical) downstream tasks \cite{kang2023benchmarking, campanella2024benchmark}. Further, Vaidya et al.\ \cite{mahmood-demographic} showed performance disparities across different demographic groups on subtyping and mutation prediction, which foundation models helped to reduce.

In contrast to performance or distribution shift evaluations, we systematically analyze to what extent hospital signatures are still encoded in foundation model representations and whether they lead to wrong predictions when correlated with prediction labels.

% Except this, to the best of our knowledge there has not been much work yet on systematically investigating to what extend foundation models reduce batch effects.

\section{Datasets and models}\label{sec:models_data}

\subsection*{Datasets}

\begin{figure}[t]
    \centering
    \includegraphics[width=1.0\linewidth]{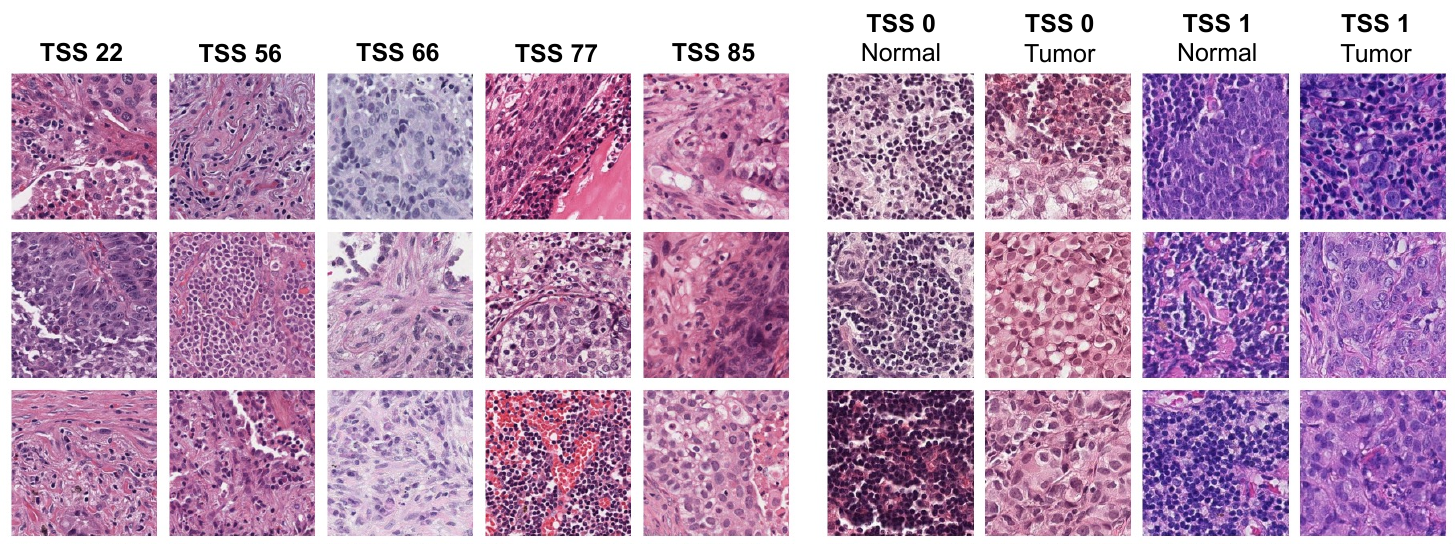}
    \caption{Arbitrarily chosen patches from the TCGA-LUSC-5 dataset (left) and the CAMELYON16 dataset (right). Differences in staining are apparent, e.g., see \textit{TSS 66}, \textit{TSS 0}, or \textit{TSS 1}.}
    \label{fig:dataset_samples}
\end{figure}

We focus on studying batch effects originating from site-specific signatures in the data, e.g., variations induced by differing cutting, staining, or scanning protocols across hospitals. We used carefully composed subsets of two public multi-site histopathology datasets for our experiments:
\begin{itemize}
    \item 
    \textbf{TCGA-LUSC-5}. The TCGA Lung Squamous Cell Carcinoma (TCGA-LUSC) project provides H\&E-stained slides containing lung squamous cell tumors \cite{Gridelli2015} from various tissue source sites. We selected five of the most contributing sites: the Mayo Clinic (\textit{TSS 22}), the International Genomics Consortium (\textit{TSS 56}), Indivumed (\textit{TSS 66}), the Prince Charles Hospital (\textit{TSS 77}), and Asterand Bioscience (\textit{TSS 85}). We sampled one diagnostic primary tumor FFPE slide ("-DX1") from up to 50 unique patients per TSS, respectively.
    \item
    \textbf{CAMELYON16}. The CAMELYON16 dataset \cite{bejnordi2017camelyon} comprises 400 sentinel lymph node slides from women with breast cancer, some of which contain precisely annotated regions of lymph node metastasis regions of different sizes. The dataset was issued by two hospitals in the Netherlands: the Radboud University Medical Center (\textit{TSS 0}) and the University Medical Center Utrecht (\textit{TSS 1}).
\end{itemize}

We extracted patches from the slides of $256 \times 256$ pixels without overlap at 20x magnification (0.5 microns per pixel). We identified and excluded background patches via Otsu's method \cite{otsu1975threshold} on slide thumbnails and applied a patch-level minimum standard deviation of 8. Each patch was labeled with the tissue source site of its slide. For CAMELYON16, we additionally labeled those patches fully inside the metastasis annotations as 'tumor' and those fully outside as 'normal'. The resulting dataset statistics and example patches are displayed in Table~\ref{tab:dataset_statistics} and Figure~\ref{fig:dataset_samples}.

We additionally aimed to assess the effect of stain normalization in combination with the foundation models, which is arguably the most popular approach intending to mitigate batch effects in histopathology (see e.g. \cite{hoque23stainnorm}). We assessed the widely used Reinhard \cite{reinhard} and Macenko \cite{macenko} techniques, and applied them to all patches in our datasets. As normalization targets, we used the average statistics of 500 patches randomly sampled from TCGA-LUSC-5. Example patches are displayed in Figure~\ref{fig:dataset_samples_norm}.

As we were primarily interested in non-informative signatures like scanning and staining characteristics, we aimed for homogeneous distributions of biological metadata (e.g., diagnosis, resection site, or stage) across tissue source sites. In both datasets, all slides come from the same organ and disease entity, respectively. Additionally, we report the statistics of available clinical metadata per tissue source sites for TCGA-LUSC-5 in Table~\ref{tab:dataset_confounders}. Even though the disease stages vary to some degree across TSS, the sites display similar distributions of clinical metadata.

\begin{table}[ht]
    \caption{Dataset statistics, i.e., the average number of patches per slide after preprocessing. For \dc{} only patches annotated as 100\% tumorous or normal are considered.}
    \label{tab:dataset_statistics}
    \centering
    \scriptsize
    \begin{tabular}{lrrrrrrrrrr}
    \toprule
    Dataset & \multicolumn{5}{c}{TCGA-LUSC-5} && \multicolumn{4}{c}{CAMELYON16} \\
    \cmidrule{2-6} \cmidrule{8-11}
     &  &  &   &  &  && TSS 0 & TSS 0 & TSS 1 & TSS 1 \\
    Subset & TSS 22 &   TSS 56 &    TSS 66 &     TSS 77 &    TSS 85 && Normal & Tumor & Normal & Tumor \\
    \midrule
    \# Slides &    37 &    41 &    38 &     44 &    50 &&    150 &    95 &     89 &    65 \\
    \# Patches per slide &  7,351 &  2,577 &  2,520 &  11,320 &  6,983 &&  6,663 &  9,090 &   9,733 &  10,715 \\
    \# Tumor patches per slide &  - &  - &  - &  - &  - && - & 508 & - & 993 \\
    % Patches (mean ± std) &  7,351 ± 2,663 &  2,577 ± 2,237 &  2,520 ± 1,396 &  11,320 ± 3,326 &  6,983 ± 3,077 & 6663 ± 4272 &  9090 ± 3981 &  9733 ± 4440 &  10715 ± 5924 \\
    % Tumor patches (mean ± std) &  - &  - &  - &  - &  - & - &  508 ± 1678 &  - & 993 ± 2696 \\
    \bottomrule
    \end{tabular}
\end{table}

\subsection*{Foundation models}

We used multiple publicly available histopathological foundation models (see Table~\ref{tab:models} for an overview). They cover a broad spectrum of architectures (convolutional vs.\ ViTs), pretraining frameworks (SimCLR, DINO, SRCL, iBOT, DINOv2), dataset diversity (6k--1,5M WSIs), and model sizes (11M--1100M parameters). 
We passed all patches from our datasets through all foundation models to obtain real-valued feature vectors from the [CLS] token representations. The patches were resized to the size expected by the respective model.

\begin{table}[ht]
    \centering
    \caption{Overview of analyzed histopathological foundation models.}
    \scriptsize
    \label{tab:models}
    \begin{tabular}{lccrcrr}
        \toprule
         Model && Pretraining objective & \#Pretraining WSIs & Architecture & \#Parameters & Feature dimension   \\
         \midrule
         Ciga &\cite{ciga2021self}  & SimCLR & (400k patches) & ResNet-18 & 11M & 512 \\
         Hipt &\cite{chen2022scaling} & DINO & 10k & ViT-S/16 & 21M & 384 \\
         RetCCL &\cite{wang2023retccl} & \makecell{cluster-guided\\contrastive} & 34k & ResNet-50 & 25M & 2048 \\
         Kang-DINO &\cite{kang2023benchmarking} & DINO & 36k & ViT-S/8 & 21M &  384 \\
         CTransPath &\cite{wang2022transformer} & SRCL  & 34k & SWIN-T & 28M & 768 \\
         Phikon &\cite{phikon} & iBOT  & 6k & ViT-B/16 & 86M & 768 \\
         UNI &\cite{chen2024uni} & DINOv2  & 100k & ViT-L/16 & 303M & 1024 \\
         ProvGigapath &\cite{xu2024gigapath} & DINOv2  & 171k &  ViT-g/14 & 1100M & 1536 \\
         Virchow &\cite{vorontsov2024virchow_natmed} & DINOv2 & 1.5M & ViT-H/14 & 631M & 1280\\
         \bottomrule
    \end{tabular}
\end{table}

\section{The role of batch effects in histopathological foundation models}\label{sec:identification}

First, we analyzed whether site signatures contained in patches are preserved in the features extracted by histopathological foundation models and how they affect downstream prediction tasks. We further assessed the effectiveness of stain normalization to mitigate such batch effects.

\subsection{Experiment 1: Source site prediction} \label{sec:source-site-prediction}

We evaluated to what extent the tissue source site of a patch can be predicted from its feature vector. We sampled roughly 50,000 patches per TSS, split them into 60\% training, 10\% validation, and 30\% test data at patient level, and trained classifiers to predict the TSS of a patch. As classifiers, we chose the nearest centroid classifier (NCC), k-nearest neighbors (KNN), and linear probing (LP). We ran this experiment for all foundation models and repeated it with the features of the stain-normalized patches. Further details are described in Appendix \ref{app:source_site_prediction}.

The results are displayed in Figure~\ref{fig:source-site-prediction}. For all foundation models, the TSS of the patches were recovered from their feature vectors with very high accuracy. An almost perfect classification was achieved via LP on CAMELYON16, and over 90\% accuracy for most foundation models on TCGA-LUSC-5. This implies that the feature representations of almost all patches carry distinct signatures of their TSS that can be identified via linear probing, regardless of the foundation model used. Even with NCC and KNN, which base their decisions on proximity of features, performances of over 90\% (CAMELYON16) and 75\% (TCGA-LUSC) were reached for most foundation models. This already indicates that these signatures rather than biological signals regularly dominate the distances in feature space. This is further explored in Section~\ref{sec:characterization}.

More recent models --- trained on more data and showing better downstream performances \cite{phikon, xu2024gigapath, chen2024uni, campanella2024benchmark} --- generally appeared to contain stronger TSS signatures. This points towards a trade-off between foundation model performance and susceptibility to batch effects. At the same time, we observed differences between recent models: NCC and KNN extracted less site information from Virchow than from UNI, ProvGigapath, or Phikon. This indicates that site signatures may be more dominant and available in the latter models, though present in all.

The stain normalization methods reduced the TSS prediction accuracy across all models, classifiers, and datasets. However, the LP accuracies were still over 80\% in TCGA-LUSC-5 and over 97\% in CAMELYON16 for most foundation models. Moreover, stain normalization seemed to have a stronger impact on NCC and KNN. We infer that Macenko and Reinhard target more dominant parts of the TSS signature, while not affecting more subtle features that can still be extracted via LP.

Therefore, despite large-scale and multi-site pretraining, histopathological foundation models do not fully eliminate batch effects, but still encode accessible TSS signatures in their feature vectors. Image-based stain normalization techniques slightly mitigate but do not remove these signatures.

\subsection{Experiment 2: Effect of site-specific signatures on downstream tasks} \label{sec:downstream-effect}
\begin{table}[b]
    \centering
    \scriptsize
    \caption{Four compositions of training data for the cancer prediction task. Each row presents the numbers of normal and cancer patches with their corresponding source site. The split identifiers (\textit{Split i -- x/y}) provide information about the varying ratios $x$ and $y$ of cancerous patches per TSS 0 and TSS 1, respectively. The allocation of test data remains fixed as 0/1. }\label{tab:TrainingSplitsCancerPrediction}
    \begin{tabular}{@{}llrrcrrcc@{}}\toprule
        && \multicolumn{2}{c}{TSS 0} & \phantom{a} & \multicolumn{2}{c}{TSS 1} & \phantom{a} & $\sum$ \\
        \cmidrule{3-4} \cmidrule{6-7}
        && \# normal & \# cancer && \# normal & \# cancer \\ \midrule
        \textit{Split 1 -- 0.5/0.5} & Training & 7,500 & 7,500 && 7,500 & 7,500 && 30,000\\
        \textit{Split 2 -- 0.67/0.33} & Training & 5,000 & 10,000 && 10,000 & 5,000 && 30,000\\
        \textit{Split 3 -- 0.83/0.17} & Training & 2,500 & 12,500 && 12,500 & 2,500 && 30,000\\
        \textit{Split 4 -- 1/0} & Training & & 15,000 && 15,000 & && 30,000\\ \midrule
        &Test & 5,000 &  &&  & 5,000 && 10,000\\
        \bottomrule
    \end{tabular}
\end{table}

\begin{figure}[t]
\centering
\includegraphics[width=\textwidth]{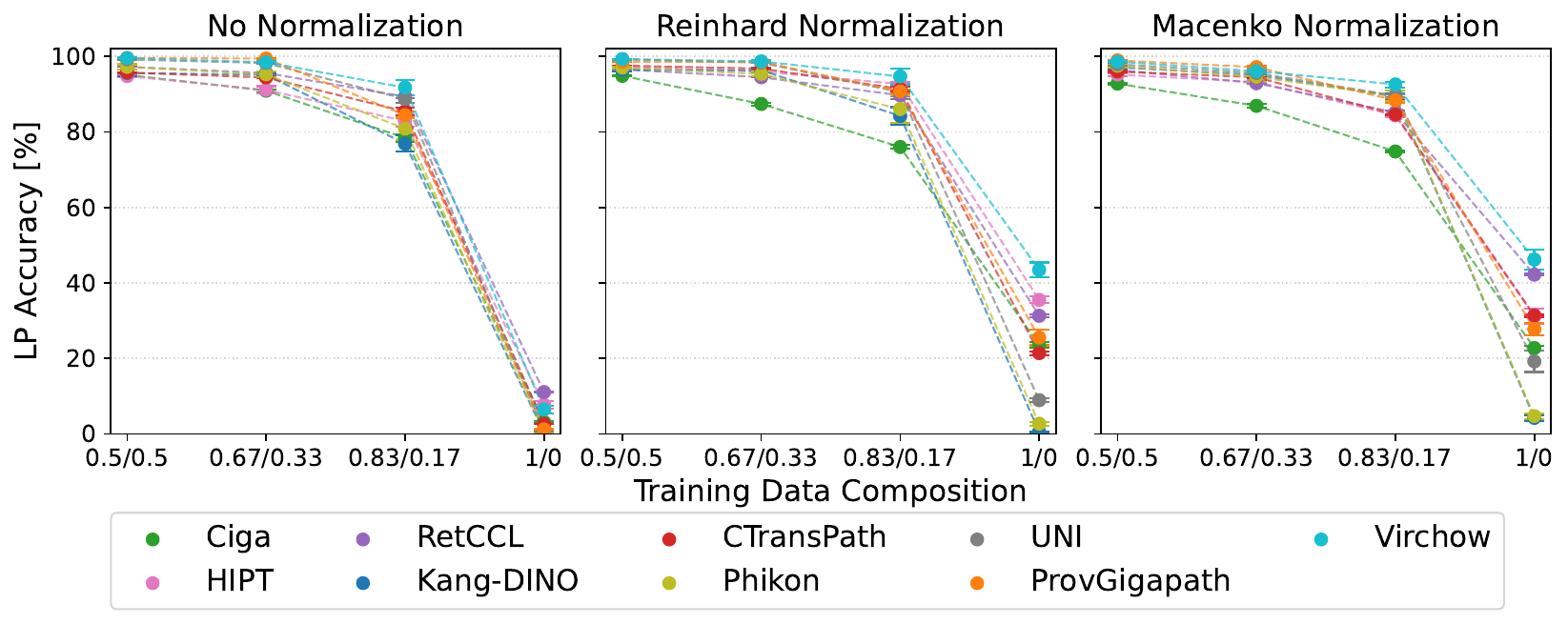}
\caption{Cancer classification accuracy in \% using \clp{} on \dc{} for different training data compositions ($x/y$). The test data composition remains 0/1 across splits. We assessed the performances for features of unnormalized, Reinhard, and Macenko normalized patches, and report mean and standard deviation over 5 repetitions.}\label{fig:CancerPrediction}
\end{figure}

We examined the impact of TSS signatures on a downstream task by creating a biased dataset with varying correlations between the classification label and the issuing TSS. The experiment is conducted on \dc{} and stipulates the classification of patches as ‘tumor’ or ‘normal’. For each TSS, we sampled 14 slides and extracted $2,500$ patches each. We created four slide-level training splits as summarized in Table~\ref{tab:TrainingSplitsCancerPrediction}. While the first split simulates a perfectly balanced data set with equally distributed tumor and normal patches, the training data of the fourth split consists of cancerous patches solely from TSS 0, whereas TSS 1 only provides healthy tissue patches. Notably, the test data, which does not change per split, is built vice versa: normal patches come from TSS 0 and cancerous ones from TSS 1. We trained LP classifiers per split and foundation model to assess whether the classifiers are affected by TSS signatures. We also repeated the experiments with the features of the stain-normalized patches (further details in Appendix \ref{app:downstream_prediction}).

The results, depicted in Figure~\ref{fig:CancerPrediction}, demonstrate that the correlation between TSS signatures and downstream task labels can lead to biased predictions and misclassifications. As the correlation increases, tumor classification accuracy drops from nearly perfect ($>90\%$ on uncorrelated data) to below random ($<20-50\%$ on fully correlated data). Significant performance drops can already be observed for the 0.83/0.17 split. This indicates that the downstream model appoints TSS information as a proxy for cancer prediction when they are in concordance.

In comparison, the accuracy of more recent models like Virchow and UNI was slightly higher and more stable than of earlier models like Ciga, particularly on the 0.67/0.33 and 0.83/0.17 splits, despite stronger encoding of TSS signatures as shown in Section~\ref{sec:source-site-prediction}. This indicates that more recent models profit from other factors aiding their generalizability, for which further investigation will be required.

Stain normalization increased the tumor prediction performance for imbalanced splits, especially on the more extreme 0.83/0.17 and 1/0 settings. However, the respective accuracies are still far below those of the balanced 0.5/0.5 split, and even below random performance for the 1/0 scenario. This shows that stain normalization has a limited impact on batch effect mitigation in feature space.

A potential explanation for our results is shortcut learning \cite{hermannfoundations, clever-hans, geirhos2020shortcut, unsupervised-clever-hans}, which occurs when models choose features not just for their predictive power but also for their availability. Hermann et al.\ \cite{hermannfoundations} observed that models prefer more available features, even when they are less predictive. For instance, in natural images, models focus on object scale, texture, or background, as these are easily accessible \cite{geirhos-texture-bias}. Translated to the histopathological domain, this might indicate that foundation models are easily misled by predominant batch effects like staining or thickness rather than biological cell alterations because of their availability, and base their predictions on those instead. Section~\ref{sec:characterization} demonstrates that TSS signatures are more available than clinical information and dominate the variance in the features, which further exacerbates shortcut learning.

In practice, many histopathological datasets suffer from batch effects, especially cohorts coming from multiple hospitals (see e.g.\ \cite{Howard2021}). Therefore, our results imply that models built on top of histopathological foundation models are not necessarily able to generalize across sites and their reported performances should be interpreted with care, as they may give correct predictions for the wrong reasons \cite{hermannfoundations, clever-hans, geirhos2020shortcut}. For training more generalizable and unbiased models, data balancing, as proposed in \cite{Howard2021}, could be applied to ensure that batch effects do not confound the prediction task. Furthermore, new pretraining strategies accounting for batch effects could lead to more robust downstream predictors.

\section{Characterization of batch effects in histopathological foundation models}\label{sec:characterization}

After establishing the presence and impact of batch effects in histopathological foundation models, we conducted multiple experiments to characterize them in greater depth. For that, we focus on four foundation models with different model sizes, dataset sizes, and pretraining frameworks: CTransPath, Phikon, UNI, and Virchow.

\subsection{Experiment 3: Visualization of distances in feature space}\label{sec:distances}

First, we analyzed how batch effects impact the distances between feature vectors of patches. We sampled a tumor reference patch from a randomly chosen cancerous slide from \dc. Subsequently, we drew $1,000$ patches from the same slide (\textit{ss}), along with $5 \times 1,000$ patches from five randomly chosen other cancerous slides from the same hospital (\textit{ossh}), and $5 \times 1,000$ patches from five cancerous slides from the other hospital (\textit{osoh}). Again, we only considered patches annotated as 100\% normal or 100\% tumor, and sampled equally from both classes for each slide.

\begin{figure}[t]
    \centering
    \includegraphics[width=\textwidth]{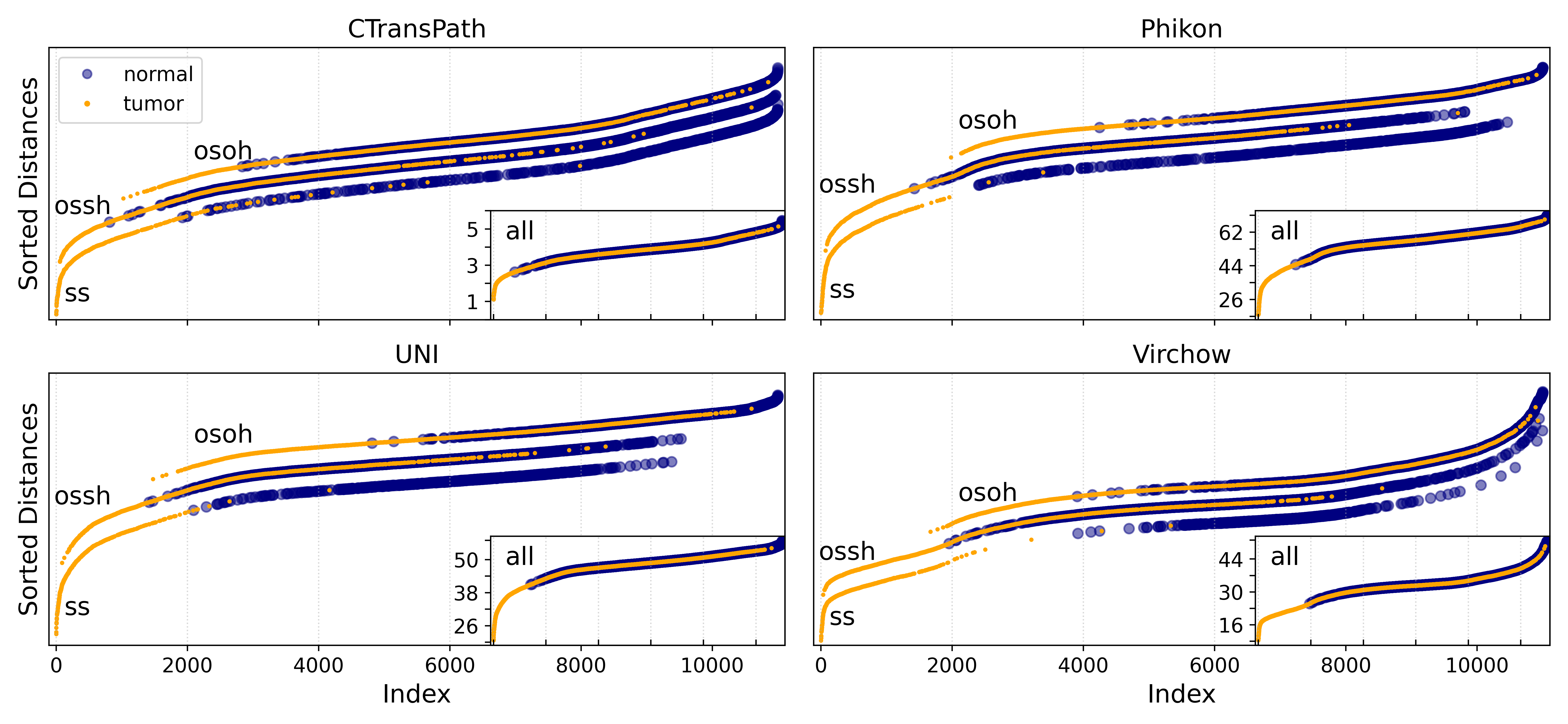}
    % ------------------------------------------------------------------------------------
    \caption{Ordered Euclidean distances between the feature of a randomly drawn cancerous reference patch and the features of \textit{ss} (bottom line), \textit{ossh} (middle line), and \textit{osoh} (top line) on \dc. Because the distances lay all on the same line (see \textit{all}), small offsets were added for clearer visualization such that \textit{ss}, \textit{ossh}, and \textit{osoh} become distinguishable.}
    \label{fig:Patches-distances-ordered}
\end{figure}

In Figure~\ref{fig:Patches-distances-ordered}, we depict the ordered Euclidean distances in the feature space to the reference patch. Most notably, there is a clear distinction between the distances to the closest patches from the same TSS compared to the ones from other TSSs. While patches from the same slide are nicely separated according to their cancer annotation, distances to normal or cancer patches largely intersect for patches from other slides. Even worse, the transition between cancer to normal patches occurs at disparate distances for \textit{ss}, \textit{ossh}, and \textit{osoh}, rendering cancerous patches of other sites as equidistant to normal patches of the same TSS.  This implies that batch effects outweigh the biological signal of cancerous or normal morphology.

In general, the experiment shows that analyses and classification methods solely based on distances between the feature vectors from histopathological foundation models are likely to fail on multi-site data due to the strong encoding of batch effects. For example, a distance-based clustering approach is unlikely to correctly identify cancer patches of different sites as belonging to the same cluster.

\subsection{Experiment 4: Site prediction performance on reduced feature data}\label{sec:reduced}

As a next step, we tried to understand how TSS signatures are encoded in the feature space of the foundation models. We specifically aimed to answer how prevalent, available, and concentrated the signatures are, and whether it is possible to isolate them. We approached these questions through the lens of dimensionality reduction and applied PCA to the feature vectors for the whole TCGA-LUSC-5 dataset per foundation model.

As a first step, we investigated to what extent site-specific information is embedded in the first principal components of the features. For that, we revisited the source site prediction task from Section~\ref{sec:source-site-prediction} using the same dataset and split. However, we trained a KNN classifier on the reduced features projected onto the first $\ell=\{1,2,3,5,10,20,30,50\}$ principal components (PCs).

\begin{figure}[t]
\centering
\includegraphics[width=\textwidth]{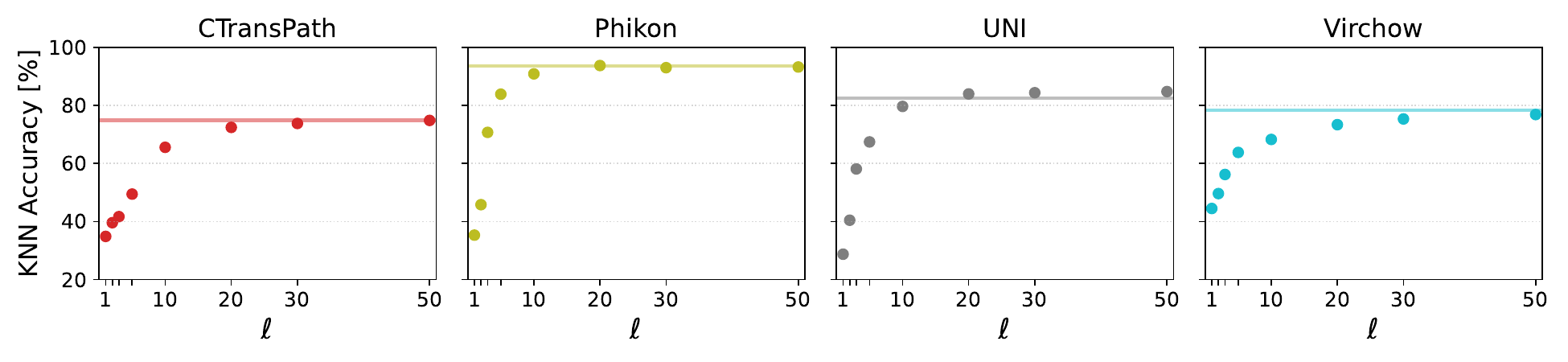}
\caption{Accuracy on site prediction task as described in Section~\ref{sec:source-site-prediction} on \dl{} using \cknn{} based on features projected onto the first $\ell$ PCs (dots) and on non-reduced features (line).}\label{fig:KnnPerformancePcaRed}
\end{figure}

The results are displayed in Figure~\ref{fig:KnnPerformancePcaRed}. For all models, high classification accuracy was already achieved with only the first few components of the data. For UNI and Phikon, we observed a steep initial increase in performance when adding principal components, almost attaining the performance on non-reduced features with only 10 PCs. For CTransPath and Virchow, this increase is less steep, still with only 20-30 PCs full feature vector performance is obtained.

This shows the high availability of TSS signatures in the feature vectors, as they are already encoded in the largest principal components. We argue that supervised downstream models are prone to utilizing these signatures as shortcuts instead of intended biological features for solving their tasks.

\subsection{Experiment 5: Component-wise separability}\label{sec:separability}

\begin{figure}[t]
\centering
\includegraphics[width=\textwidth]{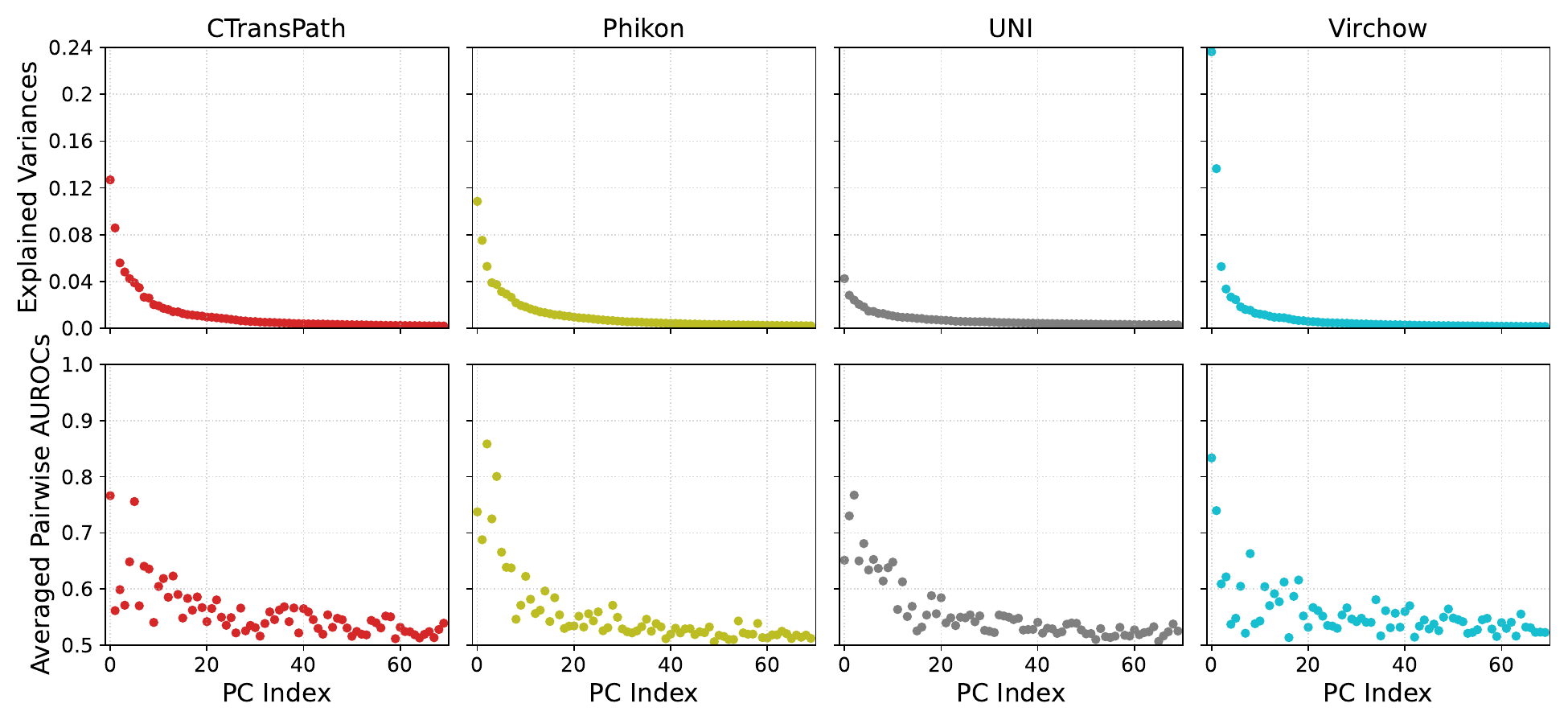}
\caption{Estimated explained variance per principal component (top) and the TSS separability of the one-dimensional feature projections on the same PCs calculated by a one-vs-one AUROC across tissue source sites (bottom) on \dl.}\label{fig:PcaComponentsAndAUROCs}
\end{figure}

While the previous experiment established that a low-dimensional representation of the features already contains sufficient information to predict their submitting site, we now examine the contribution of each principal component individually to the site-wise class separability. The features were projected onto each PC separately and we analyzed how well the projected one-dimensional features can be linearly separated. We measured their separability via a multi-class one-vs-one area under the receiver operating characteristic curve (AUROC) using the one-dimensional feature projections as scores and the TSS as labels\footnote{Instead of prediction probabilities usually required for the computation of the AUROC, we use the one-dimensional projected features as decision values. Since the orientation of these values is unknown, we compute the maximum of the one-vs-one AUROCs of both possible true label assignments. Recall that decision values above a threshold are predicted as positive so that it matters whether the reference label is set to be 0 or 1 for the computation of the ROC albeit not affecting the actual site prediction task.}. This measure naturally handles class imbalances, guarantees comparability across models as it is always in $[0.5,1]$, and is nonparametric since it does not require any further classification algorithm.

Figure~\ref{fig:PcaComponentsAndAUROCs} shows the linear class separability together with the explained variance ratio of each PC (normalized eigenvalues). We observe that many of the directions in which the data varies the most simultaneously exhibit a high site separability, making them dominant and available factors in the feature vectors. At the same time, we see that TSS signatures exist in various principal components, which makes it hard to isolate them. The pattern of how the batch effects spread across different components varies for different models.

We infer that removing site signatures is arguably difficult, as they are prevalent in various components and encoded deep within the feature space.

\section{Conclusion}\label{sec:conclusion}

In this study, we showed that histopathological foundation models do not fully eliminate batch effects. Instead, their feature vectors contain strong site-specific information, which can negatively affect the classifier performance on downstream tasks with imbalanced datasets. Image-based stain normalization methods slightly mitigate but do not remove these signatures, limiting their effect on downstream performance.
Furthermore, the site signatures dominate patch distances in feature space and are prevalent in various principal components, which makes distance and supervised learning-based methods susceptible to batch effects.

Our study comes with some limitations. The datasets we used are small, and their batch effects are known to be stark, meaning that we did not explicitly assess more subtle effects. Additionally, even though we tried sampling the data points to have similar clinical metadata across subsets, the site signatures may be partly caused by unknown biological differences for which different embeddings are desirable. We also did not identify which factors cause the batch effects to what extent, evaluate how batch effects affect generalization performance on previously unseen tissue source sites, or investigate what other factors aid or hinder the models' generalization capabilities. We leave these open questions for future work.

Regardless, we think that our results have interesting implications. First, we advocate for the assessment of batch effects to become an additional cornerstone of evaluating histopathological foundation models. Second, we think that it should be taken into account for the development of future foundation models. Third, our study opens up a promising new direction of research: developing methods for eliminating batch effects directly from feature vectors instead of tackling them on image level or during foundation model training.

\section*{Acknowledgements}

The results shown here are in whole or part based upon data generated by the TCGA Research Network: https://www.cancer.gov/tcga.

\newpage

\bibliographystyle{unsrt}
\bibliography{bibliography}

\begin{thebibliography}{10}

\bibitem{bejnordi2017camelyon}
Babak~Ehteshami Bejnordi, Mitko Veta, Paul~Johannes Van~Diest, Bram Van~Ginneken, Nico Karssemeijer, Geert Litjens, Jeroen~AWM Van Der~Laak, Meyke Hermsen, Quirine~F Manson, Maschenka Balkenhol, et~al.
\newblock Diagnostic assessment of deep learning algorithms for detection of lymph node metastases in women with breast cancer.
\newblock {\em JAMA}, 318(22):2199--2210, 2017.

\bibitem{coudray2018nsclc}
Nicolas Coudray, Paolo~Santiago Ocampo, Theodore Sakellaropoulos, Navneet Narula, Matija Snuderl, David Feny{\"o}, Andre~L Moreira, Narges Razavian, and Aristotelis Tsirigos.
\newblock Classification and mutation prediction from non--small cell lung cancer histopathology images using deep learning.
\newblock {\em Nature Medicine}, 24(10):1559--1567, 2018.

\bibitem{campanella2019clinical}
Gabriele Campanella, Matthew~G Hanna, Luke Geneslaw, Allen Miraflor, Vitor Werneck Krauss~Silva, Klaus~J Busam, Edi Brogi, Victor~E Reuter, David~S Klimstra, and Thomas~J Fuchs.
\newblock Clinical-grade computational pathology using weakly supervised deep learning on whole slide images.
\newblock {\em Nature Medicine}, 25(8):1301--1309, 2019.

\bibitem{strom2020prostate}
Peter Str{\"o}m, Kimmo Kartasalo, Henrik Olsson, Leslie Solorzano, Brett Delahunt, Daniel~M Berney, David~G Bostwick, Andrew~J Evans, David~J Grignon, Peter~A Humphrey, et~al.
\newblock Artificial intelligence for diagnosis and grading of prostate cancer in biopsies: a population-based, diagnostic study.
\newblock {\em The Lancet Oncology}, 21(2):222--232, 2020.

\bibitem{lu2021clam}
Ming~Y Lu, Drew~FK Williamson, Tiffany~Y Chen, Richard~J Chen, Matteo Barbieri, and Faisal Mahmood.
\newblock Data-efficient and weakly supervised computational pathology on whole-slide images.
\newblock {\em Nature Biomedical Engineering}, 5(6):555--570, 2021.

\bibitem{lu2021origin}
Ming~Y Lu, Tiffany~Y Chen, Drew F~K Williamson, Melissa Zhao, Maha Shady, Jana Lipkova, and Faisal Mahmood.
\newblock Ai-based pathology predicts origins for cancers of unknown primary.
\newblock {\em Nature}, 594(7861):106--110, 2021.

\bibitem{dippel2024ad}
Jonas Dippel, Niklas Prenißl, Julius Hense, Philipp Liznerski, Tobias Winterhoff, Simon Schallenberg, Marius Kloft, Oliver Buchstab, David Horst, Maximilian Alber, Lukas Ruff, Klaus-Robert Müller, and Frederick Klauschen.
\newblock Ai-based anomaly detection for clinical-grade histopathological diagnostics.
\newblock {\em NEJM AI}, 0(0):AIoa2400468, 2024.

\bibitem{kather2020pan}
Jakob~Nikolas Kather, Lara~R Heij, Heike~I Grabsch, Chiara Loeffler, Amelie Echle, Hannah~Sophie Muti, Jeremias Krause, Jan~M Niehues, Kai~AJ Sommer, Peter Bankhead, et~al.
\newblock Pan-cancer image-based detection of clinically actionable genetic alterations.
\newblock {\em Nature Cancer}, 1(8):789--799, 2020.

\bibitem{binder2021morphological}
Alexander Binder, Michael Bockmayr, Miriam H{\"{a}}gele, Stephan Wienert, Daniel Heim, Katharina Hellweg, Masaru Ishii, Albrecht Stenzinger, Andreas Hocke, Carsten Denkert, Klaus-Robert M{\"{u}}ller, and Frederick Klauschen.
\newblock Morphological and molecular breast cancer profiling through explainable machine learning.
\newblock {\em Nature Machine Intelligence}, 3(4):355--366, 2021.

\bibitem{echle2021biomarkers}
Amelie Echle, Niklas~Timon Rindtorff, Titus~Josef Brinker, Tom Luedde, Alexander~Thomas Pearson, and Jakob~Nikolas Kather.
\newblock Deep learning in cancer pathology: a new generation of clinical biomarkers.
\newblock {\em British Journal of Cancer}, 124(4):686--696, 2021.

\bibitem{arslan2024systematic}
Salim Arslan, Julian Schmidt, Cher Bass, Debapriya Mehrotra, Andre Geraldes, Shikha Singhal, Julius Hense, Xiusi Li, Pandu Raharja-Liu, Oscar Maiques, et~al.
\newblock A systematic pan-cancer study on deep learning-based prediction of multi-omic biomarkers from routine pathology images.
\newblock {\em Communications Medicine}, 4(1):48, 2024.

\bibitem{courtiol2019mesothelioma}
Pierre Courtiol, Charles Maussion, Matahi Moarii, Elodie Pronier, Samuel Pilcer, Meriem Sefta, Pierre Manceron, Sylvain Toldo, Mikhail Zaslavskiy, Nolwenn~Le Stang, Nicolas Girard, Olivier Elemento, Andrew~G Nicholson, Jean-Yves Blay, Françoise Galateau-Sallé, Gilles Wainrib, and Thomas Clozel.
\newblock Deep learning-based classification of mesothelioma improves prediction of patient outcome.
\newblock {\em Nature Medicine}, 25(10):1519--1525, 2019.

\bibitem{skrede2020outcome}
Ole-Johan Skrede, Sepp De~Raedt, Andreas Kleppe, Tarjei~S Hveem, Knut Liest{\o}l, John Maddison, Hanne~A Askautrud, Manohar Pradhan, John~Arne Nesheim, Fritz Albregtsen, et~al.
\newblock Deep learning for prediction of colorectal cancer outcome: a discovery and validation study.
\newblock {\em The Lancet}, 395(10221):350--360, 2020.

\bibitem{saillard2020survival}
Charlie Saillard, Benoit Schmauch, Oumeima Laifa, Matahi Moarii, Sylvain Toldo, Mikhail Zaslavskiy, Elodie Pronier, Alexis Laurent, Giuliana Amaddeo, H{\'e}l{\`e}ne Regnault, et~al.
\newblock Predicting survival after hepatocellular carcinoma resection using deep learning on histological slides.
\newblock {\em Hepatology}, 72(6):2000--2013, 2020.

\bibitem{chen2022multimodal}
Richard~J Chen, Ming~Y Lu, Drew~FK Williamson, Tiffany~Y Chen, Jana Lipkova, Zahra Noor, Muhammad Shaban, Maha Shady, Mane Williams, Bumjin Joo, et~al.
\newblock Pan-cancer integrative histology-genomic analysis via multimodal deep learning.
\newblock {\em Cancer Cell}, 40(8):865--878, 2022.

\bibitem{ciga2021self}
Ozan Ciga, Tony Xu, and Anne Martel.
\newblock Self supervised contrastive learning for digital histopathology.
\newblock {\em Machine Learning with Applications}, 7, 2022.

\bibitem{chen2022scaling}
Richard Chen, Chengkuan Chen, Yicong Li, Tiffany Chen, Andrew Trister, Rahul Krishnan, and Faisal Mahmood.
\newblock Scaling vision transformers to gigapixel images via hierarchical self-supervised learning.
\newblock In {\em Proceedings of the IEEE/CVF Conference on Computer Vision and Pattern Recognition (CVPR)}, 2022.

\bibitem{wang2023retccl}
Xiyue Wang, Yuexi Du, Sen Yang, Jun Zhang, Minghui Wang, Jing Zhang, Wei Yang, Junzhou Huang, and Xiao Han.
\newblock Retccl: Clustering-guided contrastive learning for whole-slide image retrieval.
\newblock {\em Medical Image Analysis}, 83:102645, 2023.

\bibitem{wang2022transformer}
Xiyue Wang, Sen Yang, Jun Zhang, Minghui Wang, Jing Zhang, Wei Yang, Junzhou Huang, and Xiao Han.
\newblock Transformer-based unsupervised contrastive learning for histopathological image classification.
\newblock {\em Medical Image Analysis}, 81:102559, 2022.

\bibitem{kang2023benchmarking}
Mingu Kang, Heon Song, Seonwook Park, Donggeun Yoo, and S{\'e}rgio Pereira.
\newblock Benchmarking self-supervised learning on diverse pathology datasets.
\newblock In {\em Proceedings of the IEEE/CVF Conference on Computer Vision and Pattern Recognition (CVPR)}, 2023.

\bibitem{phikon}
Alexandre Filiot, Ridouane Ghermi, Antoine Olivier, Paul Jacob, Lucas Fidon, Alice Kain, Charlie Saillard, and Jean-Baptiste Schiratti.
\newblock Scaling self-supervised learning for histopathology with masked image modeling.
\newblock {\em medRxiv preprint medRxiv:2023.07.21.23292757v2}, 2023.

\bibitem{kaiko2024}
Nanne Aben, Edwin~D de~Jong, Ioannis Gatopoulos, Nicolas K{\"a}nzig, Mikhail Karasikov, Axel Lagr{\'e}, Roman Moser, Joost van Doorn, Fei Tang, et~al.
\newblock Towards large-scale training of pathology foundation models.
\newblock {\em arXiv preprint arXiv:2404.15217}, 2024.

\bibitem{xu2024gigapath}
Hanwen Xu, Naoto Usuyama, Jaspreet Bagga, Sheng Zhang, Rajesh Rao, Tristan Naumann, Cliff Wong, Zelalem Gero, Javier González, Yu~Gu, et~al.
\newblock A whole-slide foundation model for digital pathology from real-world data.
\newblock {\em Nature}, 630(8015), 2024.

\bibitem{chen2024uni}
Richard~J Chen, Tong Ding, Ming~Y Lu, Drew~FK Williamson, Guillaume Jaume, Andrew~H Song, Bowen Chen, Andrew Zhang, Daniel Shao, Muhammad Shaban, et~al.
\newblock Towards a general-purpose foundation model for computational pathology.
\newblock {\em Nature Medicine}, 30(3):850--862, 2024.

\bibitem{vorontsov2024virchow_natmed}
Eugene Vorontsov, Alican Bozkurt, Adam Casson, George Shaikovski, Michal Zelechowski, Kristen Severson, Eric Zimmermann, James Hall, Neil Tenenholtz, Nicolo Fusi, et~al.
\newblock A foundation model for clinical-grade computational pathology and rare cancers detection.
\newblock {\em Nature Medicine}, 30(7):1--12, 2024.

\bibitem{Dippel2024RudolfV}
Jonas Dippel, Barbara Feulner, Tobias Winterhoff, Timo Milbich, Stephan Tietz, Simon Schallenberg, Gabriel Dernbach, Andreas Kunft, Simon Heinke, Marie-Lisa Eich, Julika Ribbat-Idel, Rosemarie Krupar, Philipp Anders, Niklas Prenißl, Philipp Jurmeister, David Horst, Lukas Ruff, Klaus-Robert Müller, Frederick Klauschen, and Maximilian Alber.
\newblock Rudolfv: A foundation model by pathologists for pathologists, 2024.

\bibitem{Howard2021}
Frederick~M Howard, James Dolezal, Sara Kochanny, Jefree Schulte, Heather Chen, Lara Heij, Dezheng Huo, Rita Nanda, Olufunmilayo~I Olopade, Jakob~N Kather, et~al.
\newblock The impact of site-specific digital histology signatures on deep learning model accuracy and bias.
\newblock {\em Nature Communications}, 12(1):4423, 2021.

\bibitem{campanella2024benchmark}
Gabriele Campanella, Shengjia Chen, Ruchika Verma, Jennifer Zeng, Aryeh Stock, Matt Croken, Brandon Veremis, Abdulkadir Elmas, Kuan-lin Huang, Ricky Kwan, et~al.
\newblock A clinical benchmark of public self-supervised pathology foundation models.
\newblock {\em arXiv preprint arXiv:2407.06508}, 2024.

\bibitem{reinhard}
Erik Reinhard, Michael Adhikhmin, Bruce Gooch, and Peter Shirley.
\newblock Color transfer between images.
\newblock {\em IEEE Computer Graphics and Applications}, 21(5):34--41, 2001.

\bibitem{macenko}
Marc Macenko, Marc Niethammer, J.~Marron, David Borland, John Woosley, Xiaojun Guan, Charles Schmitt, and Nancy Thomas.
\newblock A method for normalizing histology slides for quantitative analysis.
\newblock In {\em Proceedings of the IEEE International Symposium on Biomedical Imaging: From Nano to Macro (ISBI)}, 2009.

\bibitem{histo-xai-review}
Frederick Klauschen, Jonas Dippel, Philipp Keyl, Philipp Jurmeister, Michael Bockmayr, Andreas Mock, Oliver Buchstab, Maximilian Alber, Lukas Ruff, Gr\'{e}goire Montavon, and Klaus-Robert M\"{u}ller.
\newblock Toward explainable artificial intelligence for precision pathology.
\newblock {\em Annual Review of Pathology: Mechanisms of Disease}, 19(1):541--570, 2024.

\bibitem{clever-hans}
Sebastian Lapuschkin, Stephan W{\"a}ldchen, Alexander Binder, Gr{\'e}goire Montavon, Wojciech Samek, and Klaus-Robert M{\"u}ller.
\newblock Unmasking clever hans predictors and assessing what machines really learn.
\newblock {\em Nature Communications}, 10(1):1096, 2019.

\bibitem{geirhos2020shortcut}
Robert Geirhos, J{\"o}rn-Henrik Jacobsen, Claudio Michaelis, Richard Zemel, Wieland Brendel, Matthias Bethge, and Felix~A Wichmann.
\newblock Shortcut learning in deep neural networks.
\newblock {\em Nature Machine Intelligence}, 2(11):665--673, 2020.

\bibitem{hermannfoundations}
Katherine Hermann, Hossein Mobahi, Thomas Fel, and Michael~C. Mozer.
\newblock On the foundations of shortcut learning.
\newblock In {\em International Conference on Learning Representations (ICLR)}, 2024.

\bibitem{unsupervised-clever-hans}
Jacob Kauffmann, Jonas Dippel, Lukas Ruff, Wojciech Samek, Klaus-Robert M{\"u}ller, and Gr{\'e}goire Montavon.
\newblock The clever hans effect in unsupervised learning.
\newblock {\em arXiv preprint arXiv:2408.08041}, 2024.

\bibitem{navarro2022self}
Fernando Navarro, Christopher Watanabe, Suprosanna Shit, Anjany Sekuboyina, Jan~C Peeken, Stephanie~E Combs, and Bjoern~H Menze.
\newblock Self-supervised pretext tasks in model robustness \& generalizability: A revisit from medical imaging perspective.
\newblock In {\em 44th Annual International Conference of the IEEE Engineering in Medicine \& Biology Society (EMBC)}, 2022.

\bibitem{goyal-ssl-robust}
Priya Goyal, Quentin Duval, Isaac Seessel, Mathilde Caron, Ishan Misra, Levent Sagun, Armand Joulin, and Piotr Bojanowski.
\newblock Vision models are more robust and fair when pretrained on uncurated images without supervision.
\newblock {\em arXiv preprint arXiv:2202.08360}, 2022.

\bibitem{mahmood-demographic}
Anurag Vaidya, Richard~J Chen, Drew~FK Williamson, Andrew~H Song, Guillaume Jaume, Yuzhe Yang, Thomas Hartvigsen, Emma~C Dyer, Ming~Y Lu, Jana Lipkova, et~al.
\newblock Demographic bias in misdiagnosis by computational pathology models.
\newblock {\em Nature Medicine}, 30(4):1174--1190, 2024.

\bibitem{Gridelli2015}
Cesare Gridelli, Antonio Rossi, David Carbone, Juliana Guarize, Niki Karachaliou, Tony Mok, Francesco Petrella, Lorenzo Spaggiari, and Rafael Rosell.
\newblock Non-small-cell lung cancer.
\newblock {\em Nature Reviews Disease Primers}, 1:15009, 2015.

\bibitem{otsu1975threshold}
Nobuyuki Otsu.
\newblock A threshold selection method from gray-level histograms.
\newblock {\em IEEE Transactions on Systems, Man, and Cybernetics (SMC)}, 9(1):62--66, 1979.

\bibitem{hoque23stainnorm}
Ziaul Hoque, Anja Keskinarkaus, Pia Nyberg, and Tapio Seppänen.
\newblock Stain normalization methods for histopathology image analysis: A comprehensive review and experimental comparison.
\newblock {\em Information Fusion}, 102:101997, 2024.

\bibitem{geirhos-texture-bias}
Robert Geirhos, Patricia Rubisch, Claudio Michaelis, Matthias Bethge, Felix~A. Wichmann, and Wieland Brendel.
\newblock Imagenet-trained {CNNs} are biased towards texture; increasing shape bias improves accuracy and robustness.
\newblock In {\em 7th International Conference on Learning Representations (ICLR)}, 2019.

\end{thebibliography}

\newpage

%%%%%%%%%%%%%%%%%%%%%%%%%%%%%%%%%%%%%%%%%%%%%%%%%%%%%%%%%%%%

\appendix

\section{Appendix}

\subsection{Dataset details}

\begin{figure}[h!]
    \centering
    \includegraphics[width=1.0\linewidth]{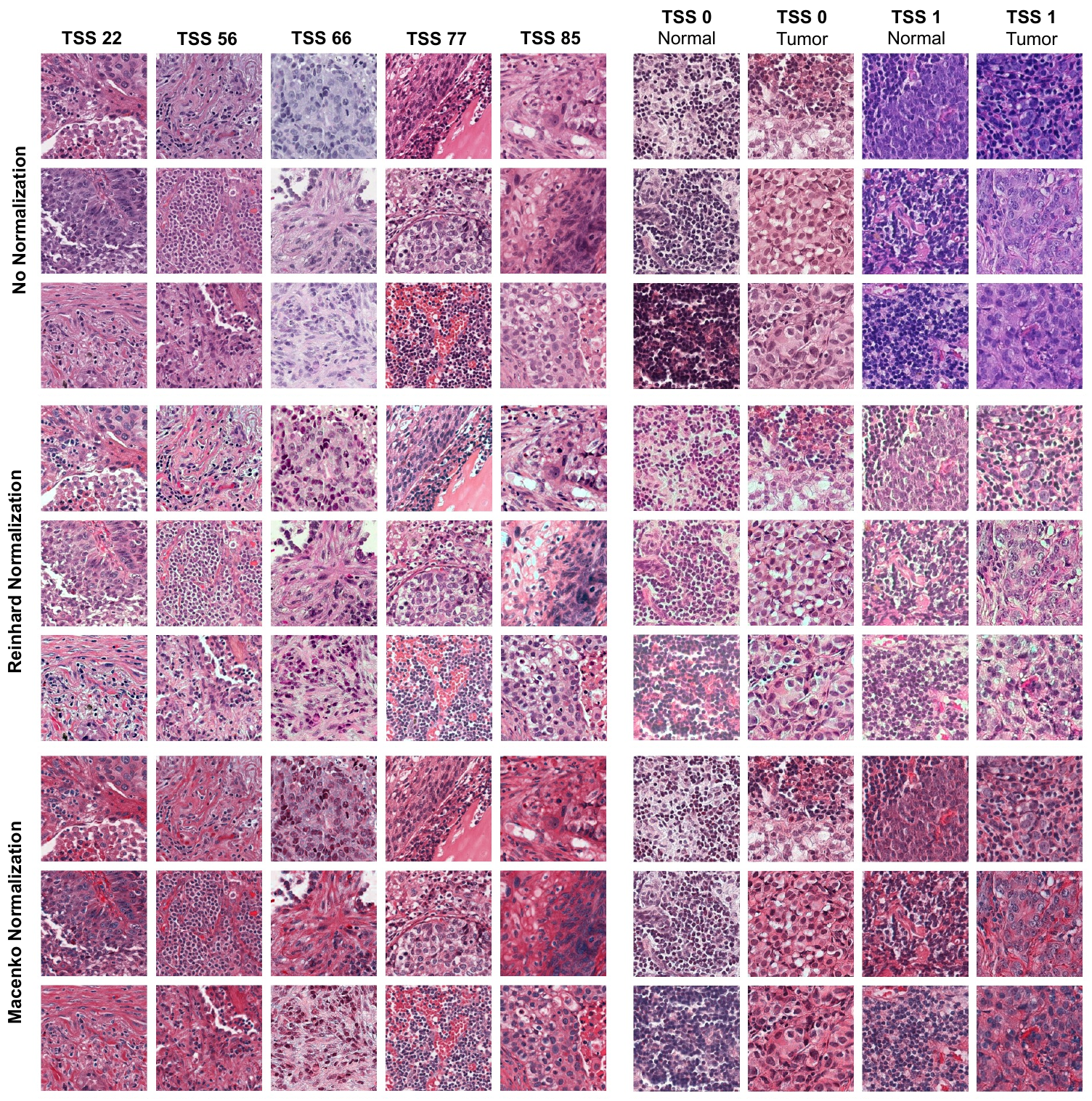}
    \caption{Arbitrarily chosen patches from the TCGA-LUSC-5 dataset (left) and the CAMELYON16 dataset (right) with Reinhard \cite{reinhard} and Macenko \cite{macenko} stain normalization.}
    \label{fig:dataset_samples_norm}
\end{figure}

Example patches and metadata statistics of TCGA-LUSC-5 are provided in Figure~\ref{fig:dataset_samples_norm} and Table~\ref{tab:dataset_confounders}, respectively.

\begin{table}[t]
    \caption{Selected metadata statistics of the TCGA-LUSC-5 dataset. We chose \textit{primary\_diagnosis}, \textit{site\_of\_resection\_or\_biopsy}, \textit{ajcc\_pathologic\_t}, and \textit{ajcc\_pathologic\_stage} as possibly relevant columns in the clinical metadata of the TCGA LUSC project. We report the number of slides per stage and subset. SCC = squamous cell carcinoma; NOS = not otherwise specified.}
    \label{tab:dataset_confounders}
    \centering
    \scriptsize
    \begin{tabular}{lrrrrr}
    &&&&& \\
    \toprule
    Primary diagnosis & TSS 22 & TSS 56 & TSS 66 & TSS 77 &  TSS 85 \\
    \midrule
    Papillary SCC                  &   0 &   0 &   0 &   1 &   2 \\
    SCC, NOS                       &  36 &  39 &  38 &  40 &  45 \\
    SCC, keratinizing, NOS         &   0 &   1 &   0 &   3 &   2 \\
    SCC, large cell, nonkeratinizing, NOS &   1 &   1 &   0 &   0 &   1 \\
    \midrule
    &&&&& \\
    \midrule
    Site of resection or biopsy & TSS 22 & TSS 56 & TSS 66 & TSS 77 &  TSS 85 \\
    \midrule
    Lower lobe, lung            &  13 &  16 &  16 &  12 &  13 \\
    Lung, NOS                   &   9 &   0 &   0 &   8 &   0 \\
    Main bronchus               &   0 &   0 &   0 &   1 &   0 \\
    Middle lobe, lung           &   0 &   0 &   3 &   1 &   2 \\
    Overlapping lesion of lung  &   1 &   1 &   0 &   1 &   0 \\
    Upper lobe, lung            &  14 &  24 &  19 &  21 &  35 \\
    \midrule
    &&&&& \\
    \midrule
    AJCC T stage & TSS 22 & TSS 56 & TSS 66 & TSS 77 &  TSS 85 \\
    \midrule
    T1                &   6 &   0 &   4 &   5 &   2 \\
    T1a               &   3 &   3 &   0 &   0 &   6 \\
    T1b               &   4 &   6 &   0 &   0 &  10 \\
    T2                &   9 &   3 &  28 &  23 &   1 \\
    T2a               &   8 &  14 &   0 &   6 &  13 \\
    T2b               &   2 &   7 &   0 &   0 &   8 \\
    T3                &   4 &   8 &   0 &   7 &   9 \\
    T4                &   1 &   0 &   6 &   3 &   1 \\
    \midrule
    &&&&& \\
    \midrule
    AJCC stage & TSS 22 & TSS 56 & TSS 66 & TSS 77 &  TSS 85 \\
    \midrule
    '--                   &   1 &   0 &   0 &   0 &   0 \\
    Stage IA              &  10 &   8 &   3 &   3 &  13 \\
    Stage IB              &  15 &  15 &  14 &  14 &  11 \\
    Stage IIA             &   5 &   8 &   0 &   4 &  12 \\
    Stage IIB             &   1 &   6 &   6 &  13 &   9 \\
    Stage IIIA            &   5 &   4 &   3 &   7 &   5 \\
    Stage IIIB            &   0 &   0 &  11 &   3 &   0 \\
    Stage IV              &   0 &   0 &   1 &   0 &   0 \\
    \bottomrule
    \end{tabular}
\end{table}

\subsection{Experimental details}

\subsubsection{Experiment 1: Source site prediction} \label{app:source_site_prediction}

For both datasets, we randomly selected $n_i = \frac{50,000}{m_i}$ patches per slide of TSS $i$ with $m_i$ being the number of slides of site $i$. If a slide contained less than $n_i$ patches, we drew all the available patches. This resulted in approximately 250,000 patches for TCGA-LUSC-5 and 100,000 for CAMELYON16 with roughly balanced site labels. We split each dataset into 60\% training, 10\% validation, and 30\% test data at patient level, ensuring that no patches from the same slide appear in both training and validation or test set.

We used three classifiers of increasing expressivity to predict the source site of the patches' feature vectors: nearest centroid classifier (NCC), k-nearest neighbors (KNN), and linear probing (LP). For KNN, we used the $k=5$ nearest neighbors from the train set to predict the source site of a test patch. For LP, we used the Adam optimizer, an initial learning rate of 0.001, and the cross-entropy loss with a batch size of 128 to train the single-layer classification head. We trained for 20 epochs and selected the model with the lowest validation loss, which we then applied to the held-out test set. We computed the accuracy score by comparing the class with the highest predicted logit with the sample label.

\subsubsection{Effect of site-specific signatures on downstream tasks} \label{app:downstream_prediction}

For each TSS, we sampled 7 normal and 7 tumor slides (total: 28 slides), ensuring that each contained at least 2,500 patches that are 100\% normal or 100\% tumor according to the pathologist annotation, respectively. For each normal slide, we then sampled 2,500 patches that are 100\% normal, and for each tumor slide 2,500 patches that are 100\% tumorous. From this pool of patches, we created splits as described in Table~\ref{tab:TrainingSplitsCancerPrediction}. All splits were created on slide level, and all 2,500 patches from a selected slide were used: for instance, if Split 2 contains 10,000 cancer patches from TSS 0, it means that they come from 4 different tumor slides contributing 2,500 distinct tumor patches, each. The patches used for the test set were the same across different training splits. We also added 5,000 patches from different slides as validation sets, having the same TSS-label compositions as the training splits, respectively. We highlight that in any case, patches from the same slide only appeared in either train, validation, or test set, but never more than one of the subsets.

We applied linear probing (LP) to predict whether a patch is cancerous or normal. We used the Adam optimizer, an initial learning rate of 0.001, and the cross-entropy loss with a batch size of 128 to train the single-layer classification head. We trained for 20 epochs and selected the model with the lowest validation loss, which we then applied to the held-out test set. We computed the accuracy score by comparing the class with the highest predicted logit with the sample label. We repeated the training process 5 times on the same splits, and report the mean and standard deviation of the accuracy scores.

\subsubsection{Compute resources} \label{app:compute}

Data processing and feature extraction with the foundation models were conducted on our internal cluster equipped with GPUs having at least 24GB GPU memory, each. The datasets required roughly 1TB of disk memory. As the extracted feature vectors only made up a small fraction of the total memory used, all experiments could be executed on a standard CPU.

%%%%%%%%%%%%%%%%%%%%%%%%%%%%%%%%%%%%%%%%%%%%%%%%%%%%%%%%%%%%

\end{document}